\documentclass{article}
\usepackage[utf8]{inputenc}
\usepackage[english]{babel}
\usepackage{todonotes}
\usepackage{ulem}
\usepackage{csquotes}
\usepackage{url}
\usepackage{pdfpages}
\usepackage[title]{appendix}
\usepackage{hyperref}

\usepackage{array}
\usepackage{boldline}

\usepackage{natbib}
\setcitestyle{authoryear,open={(},close={)}}

\newcommand{\formRQ}{\textbf{A.1}}

\newcommand{\formDV}{\textbf{A.2}}

\newcommand{\formIV}{\textbf{A.3}}

\newcommand{\formTrainData}{\textbf{A.4}}

\newcommand{\formTestData}{\textbf{A.5}}

\newcommand{\formFiltering}{\textbf{A.6}}

\newcommand{\formMetrics}{\textbf{A.7}}

\newcommand{\formBaselines}{\textbf{A.8}}

\newcommand{\formOtherA}{\textbf{A.9}}

\newcommand{\formPredMethod}{\textbf{B.1}}

\newcommand{\formModelTrain}{\textbf{B.2}}

\newcommand{\formTestAccess}{\textbf{B.3}}

\newcommand{\formOther}{\textbf{B.4}}

\newcommand{\formAdjustments}{\textbf{B.5}}

\newcommand*\samethanks[1][\value{footnote}]{\footnotemark[#1]}

\usepackage{comment}
\usepackage{enumitem}
\usepackage{geometry}
\title{Pre-registration for Predictive Modeling}

\author{ 
Jake M. Hofman \thanks{Microsoft Research} \and
Angelos Chatzimparmpas \thanks{Northwestern University} \and
Amit Sharma \samethanks[1] \and
Duncan J. Watts \thanks{University of Pennsylvania} \and
Jessica Hullman \samethanks[2]
}  

\date{\today}

\begin{document}

\maketitle
\begin{abstract}

Amid rising concerns of reproducibility and generalizability in predictive modeling, we explore the possibility and potential benefits of introducing pre-registration to the field. Despite notable advancements in predictive modeling, spanning core machine learning tasks to various scientific applications, challenges such as overlooked contextual factors, data-dependent decision-making, and unintentional re-use of test data have raised questions about the integrity of results. To address these issues, we propose adapting pre-registration practices from explanatory modeling to predictive modeling. We discuss current best practices in predictive modeling and their limitations, introduce a lightweight pre-registration template, and present a qualitative study with machine learning researchers to gain insight into the effectiveness of pre-registration in preventing biased estimates and promoting more reliable research outcomes. We conclude by exploring the scope of problems that pre-registration can address in predictive modeling and acknowledging its limitations within this context.

\end{abstract}

\section{Introduction}

Several scientific communities are currently facing a replication crisis, wherein it has proven difficult or impossible for researchers to independently verify the results of previously published studies.
Failures to replicate large swaths of experimental work~\citep{camerer2018evaluating,nosek2015,begley2012raise,baker20161} have come in fields like psychology or medicine, that focus on what \cite{hofman2021integrating} call \textit{explanatory modeling}, where the goal is to identify and estimate causal effects (e.g., is there an effect of X on Y, and if so, how large is it?).
While there are many different factors that can contribute to unreliable findings in explanatory modeling, the combination of small-scale experiments involving noisy measurements and the (mis)use of null hypothesis significance testing (NHST) has received a great deal of attention in recent years.
Under these conditions, researchers can mistake idiosyncratic patterns in noise for true effects, resulting in unreliable findings that do not replicate upon further investigation~\citep{button2013power,loken2017measurement,meehl1990summaries,simmons2011false}.
More generally, some forms of data-dependent decision making (e.g., about how to define research questions or hypotheses, how to filter or transform data, how to model data, what tests to run, etc.) can lead to similar problems regardless of the specifics of the methods ~\citep{gelman2013garden}.

What about other fields, such as machine learning and data science, that focus
less on explanation and more on \textit{predictive modeling}, defined in \cite{hofman2021integrating} as directly forecasting outcomes (e.g., how well can an outcome Y be predicted using all available features X?) without necessarily focusing on isolating individual causal effects? Predictive modeling is typically done by testing (out-of-sample) predictions on large-scale datasets, and hence---unlike explanatory modeling---involves neither small experiments nor misuse of significance testing.
With advances in the fields of statistics and machine learning (ML) we have seen remarkable performance gains in predictive modeling over the last decade, for both traditional ML tasks and for scientific applications. The same methods that have been shown to achieve at or above human-level performance on tasks like playing chess, classifying images, or understanding natural language~\citep{zhang2021ai,willingham2023,bubeck2023sparks,norvig2023agi} have also been used for scientific inquiry and discovery in fields ranging from quantitative social science to biology to physics~\citep{salganik2020measuring,libbrecht2015machine,carleo2019machine}.
Given this rapid progress, it may seem that predictive modeling---whether deployed for routine ML tasks or for scientific applications---is immune to the replication issues that have plagued explanatory modeling.

Here, we caution against this conclusion.
In fact, at the same time that predictive modeling has seen great progress, there have also been growing concerns about the replicability~\citep{kapoor2021irreproducible,mcdermott2021reproducibility,pineau2021improving} 
of both core and applied ML research.
These concerns include oversights in documenting ``contextual factors'' (e.g., random seeds, computational budgets, or hyperparameters) that, if altered, can contribute sizable variance to results; data-dependent decisions about key aspects of a modeling problem that can make it seem artificially easy; and (unintentional) re-use of test data that generate overly optimistic results.
Thus, while predictive modeling may not suffer from issues like small datasets and abuse of hypothesis testing that affect explanatory modeling, there are still many ``researcher degrees of freedom'' that can affect how problems are operationalized and how models are developed and tested, resulting in replication issues~\citep{hullman2022worst}.

In the spirit of integrative modeling~\citep{hofman2021integrating}, we propose adapting the practice of pre-registration from the explanatory modeling community~\citep{wagenmakers2012agenda,nosek2018preregistration,simmons2021pre} in an attempt to mitigate replication and generalization issues in predictive modeling.
We first discuss the status quo for best practices in predictive modeling, and then several ways in which these best practices can lead to brittle published results.
We then review how pre-registration has been used for explanatory modeling, and provide a lightweight template for scientists to use when pre-registering predictive studies.
We next present a qualitative study with in-depth interviews of practicing machine learning researchers who used the template for an example prediction exercise. Study participants noted important strengths of pre-registration for predictive modeling research, and multiple participants might have produced biased estimates of out-of-sample performance had they not adhered to the protocol.
Finally, we discuss which existing problems pre-registration can address if used correctly, and which problems it may help with but where there are fewer guarantees based on the nature of the problem.

\section{The status quo}

The last several decades have seen an explosion in predictive modeling owing to advances in fields like statistics and machine learning. %
Because explaining how a model makes predictions is not usually a primary goal in predictive modeling, as techniques have advanced so too has the complexity of models considered and the size of the datasets required to fit them. These days it is not uncommon to see models with millions or billions of parameters, and datasets with a similar number of observations. 

When dealing with such complex models, it is well known that one can encounter the problem of overfitting, 
where a model erroneously picks up on noise (at the expense of signal) in the set of examples it is fit to.  
This is similar in spirit to the problems faced in social sciences when researchers manually test many different hypotheses on a relatively small experimental dataset, but on a larger, algorithmic scale. 
In both settings, considering enough different explanations for a given dataset can lead researchers to find a model or hypothesis that appears to explain these observations, but that fails to generalize to new cases.

The field of machine learning has developed best practices aimed at avoiding this sort of overfitting. The core idea is relatively simple: instead of judging a model by how well it fits the examples it was trained on, examine its performance on a new, independent set of examples drawn from the same data-generating process. This out-of-sample testing focuses the modeling process on generalizing to new data, and is now considered standard in fields that employ predictive models.

\begin{figure}
    \centering
    \includegraphics[width=\textwidth]{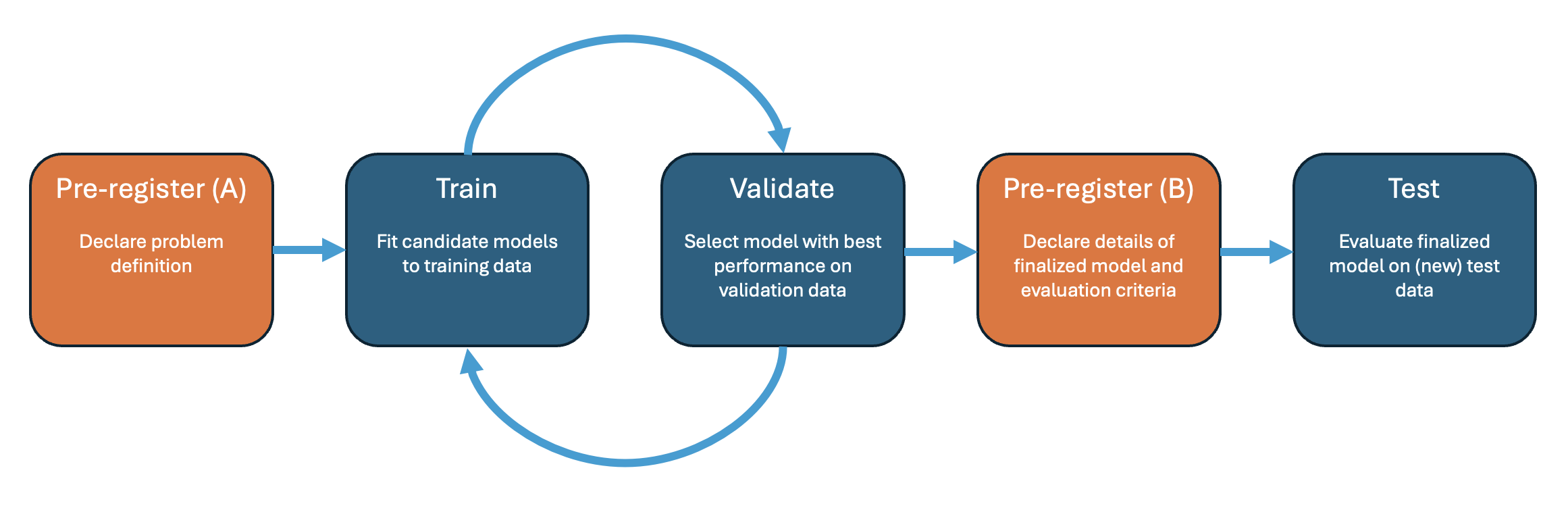}
    \caption{A diagram showing the suggested flow for incorporating pre-registration into predictive modeling. The blue boxes show the current flow without pre-registration of a train/validate/test split. The orange boxes show the suggested addition of a two-step pre-registration process, with the first step prior to training and the second step prior to testing.}
    \label{fig:prereg-flow-diagram}
\end{figure}

This procedure, often implemented through a process known as cross-validation, is summarized by the blue boxes in Figure~\ref{fig:prereg-flow-diagram}. First one considers a set of candidate models (e.g., in the case of fitting a polynomial in one variable, one candidate would be a linear model, another a quadratic model, and so on) and fits each model to the data instances in the \textit{training set} (determining in this case the best-fit coefficients for the linear model, the quadratic model, etc.). Then the quality of each fitted model is assessed on a \textit{validation set} consisting of a new set of examples that are separate from the data the models were fit to. The candidate with the best performance on the validation data is chosen as the final model, and its performance is assessed on the \textit{test set}, a third set of examples that are again separate from the data used in the previous two steps. The reason for the test set is that while one is free to iterate between the training and validation steps to develop and refine the set of models considered in these stages, this process could lead one to overfit to the validation data if enough iterations are done. The test set, however, is meant to be used once and only once, to provide an honest assessment of how well the chosen model will perform on data it (and the modeler) have never seen before.

Ideally, the researcher has access to a continuous stream of new data for each of the training, validation, and testing steps. However, in practice, researchers often have one static dataset and synthetically create different splits of it, for instance by shuffling the data and partitioning it into separate train, validation, and test datasets.
In cases where there is consensus on the details of a particular prediction problem, the common task framework can be used to improve upon researchers creating their own train/test splits~\cite{bennett2007netflix,russakovsky2015imagenet,donoho201850}. In this paradigm, an organizer posts a public dataset for training and validation, and participating teams or individuals work independently on solving the problem with it. When teams arrive at a solution, they can submit predictions for test examples, but the ground truth outcomes for the test set are kept secret from participants to measure generalization error while preventing overfitting.
\section{Problems}

In theory, evaluating a predictive model with the train/validate/test paradigm should provide a reliable estimate of how well it will generalize to new data. However, in practice, studies that report using this approach can produce misleading results for a variety of reasons, some relatively straightforward and some more complex.

One of the simplest reasons for unreliable results is the (often unintentional) failure to adequately separate the training and validation sets from the test set, known as ``leakage''. 
Sometimes this is as simple as mistakenly including the test examples in the training process~\citep{wang2019comparing,oner2020training}.
Other times
there are more subtle forms of ``leakage''.
One potentially common form is due to continuing to optimize a model after having evaluated it on the test set
(e.g., by adjusting hyperparameters or changing evaluation metrics after accessing the test set), known as ``over-hyping''~\citep{hosseini2020tried}.
While multiple accesses to the test set clearly violate the underlying principle of out-of-sample testing, in practice it may be common for practitioners to accidentally make this mistake in the same way that ``HARK-ing'', or hypothesizing after results are known, is common in explanatory modeling~\citep{kerr1998harking}.
Less obvious sources of leakage can occur due to pre-processing or feature selection applied to both training and test data, use of illegitimate features that leak information about the outcome variable, lack of independence between training and test data, and use of future data in time series modeling, among others~\citep{kapoor2022leakage}.

Performance estimates may also be brittle due to contextual factors (such as computational budgets, random seeds, and particular algorithm implementations) that go unreported despite potentially having large impacts on model generalization~\citep{dodge2019show}. Similarly, the reported benefits of new models may be inflated because researchers may invest more attention to tuning their proposed approach and less to baselines against which they compare. %

The above issues are further complicated by the fact that the train/validate/test paradigm is designed to estimate out-of-sample but in-distribution performance, assuming independent and identically distributed (I.I.D.) examples.
But even in cases where the train/validate/test paradigm is correctly applied without any forms of leakage, with all contextual factors reported, and for a problem with IID examples, there is a broader concern about potential differences in how a problem is framed and how it is \textit{operationalized}, akin to issues of construct validity in the social sciences~\citep{cronbach1955construct}.
This can be especially problematic in cases where there are many degrees of freedom in how a high-level research question is translated into a specific predictive modeling exercise.

To make this concrete, consider the many choices that define a learning problem and modeling pipeline. These include what outcome to predict; how to define training, validation, and test data; how success is measured and reported via specific metrics; how data are filtered and transformed; details of model training (e.g., random seeds, hyperparameters, etc.);  and what baseline models or methods are compared to.
Whenever these decisions are not determined in advance of viewing performance estimates, there is a risk that researchers will make these choices in ways that improve their ability to obtain satisfying results in light of their research goals. 

For instance, imagine a researcher interested in the extent to which depression in teens can be predicted from available data on teens' attitudes or behaviors~\citep{orben2019association}. 
They obtain the results of a large survey of high schoolers~\citep{Miech2000Monitoring} containing items that might be associated with depression, such as Likert-style ratings of how generally happy versus hopeless the student feels on a daily basis. 
As they begin training the model, they face choices of: which dependent variable to use, how to treat the problem (e.g., classification versus regression), how to transform the outcome variable (e.g., to construct a binary outcome variable from a Likert-style rating), how to select and transform features, how to deal with missing data, which metric they will report to evaluate test set performance (e.g., $R^2$, root mean squared error, mean absolute error), and how to define any baseline algorithms they will compare their results to.

Imagine for example that the researcher wishes to demonstrate that depression can be predicted in these data. 
By exploring possible dependent variables (i.e., survey items that might be interpreted as depression scales) and subsets of the data (e.g., by grade level), they identify a combination that appears to maximize their preferred metric.
Or, perhaps they expect depression to be related to drug use, based on previous findings from smaller studies in the literature, and want to demonstrate this on new data.
They might be predisposed to choose both the outcome variable and the metric they report conditional on what pair shows the largest difference between the performance of a model that includes versus does not include the drug-related survey items as predictors. 
Whenever the goals of the work are to support a pre-determined hypothesis (e.g., ``It is possible to predict event Y from dataset X'', ``Training technique Z improves performance on benchmark B", etc.) there may be a temptation to do the predictive equivalent of “p-hacking”, in that one tends to report outcomes that look good without thinking critically about whether or not they answer the substantive question at hand.
Importantly, even when such exploration is conducted \textit{only during model training and validation}, the results are at risk of being overly optimistic as a result of the learning problem having been cherry-picked to maximize performance.

After the researcher has trained the model and proceeds to calculate test set performance, it is critical that they do not return to the training and validation phase or otherwise continue to manipulate the model. Imagine that upon applying their model to the test data, the researcher predicting depression observes a lower-than-expected accuracy in light of the validation accuracy. They might consider ways in which their model overfit the training data. In an effort to address the overfitting, they might try various modifications, such as adding a regularizing term to the loss function and retraining. 
The (presumably higher) test accuracy they obtain after this adjustment is no longer an unbiased estimate of out-of-sample performance. %

Such concerns about the flexibility afforded by different modeling choices are far from hypothetical.
Take, for example, 
\citet{hofman2017prediction}, which looks at the problem of predicting the size of diffusion cascades on social media (i.e., how many re-posts a post will get) based on properties of the seed (i.e., the person who created the original post).
In this case, even with the dataset and model specification held fixed, exercising just a few degrees of freedom (specifically, treating the problem as classification versus regression, filtering or thresholding data, and choosing different performance metrics) is shown to change the qualitative nature of the results.
For instance, if one operationalizes this exercise as a classification problem by building a model to predict whether a post will receive at least 10 re-posts, one can obtain impressive-sounding accuracies close to 100\%. 
If, however, one treats this as a regression exercise to predict how many re-posts a given post gets without any data filtering, the $R^2$ hovers around a relatively modest 35\%.
The reason for the difference comes from the fact that only a small fraction of posts exceed the threshold of 10 re-posts, and predicting which posts do---and how far they spread---is quite challenging.
Given the stark difference in these numbers and pressure to report positive, strong results, a researcher who tries both approaches, even if only during the training and validation phase, might, upon seeing the results, publish the classification model as a success and discard the regression model as a failure.

Taken together, these issues emphasize two key points.
First, the importance of carefully executing and documenting the details of the train/validate/test process, along with an awareness of its limitations.
And second, the need for thoughtful consideration about how the prediction problem one solves relates to the higher-level goals and motivation of the modeling exercise.

\section{A pre-registration protocol for predictive modeling} \label{sec:protocol}

The solution we propose is to adapt the procedure of pre-registration to predictive modeling.
Pre-registration amounts to declaring an analysis plan before conducting it~\citep{wagenmakers2012agenda,nosek2018preregistration,simmons2021pre}. 
In experimental science, platforms like OSF\footnote{\url{https://help.osf.io/article/158-create-a-preregistration}} and AsPredicted\footnote{\url{https://aspredicted.org}} provide forms that ask the researcher to specify aspects of their research design and analysis, such as hypotheses, sampling plans, data exclusions, outcome measures, statistical models, and key inferential criteria.
Upon submission of a plan, both platforms generate a time-stamped document which the author can choose to make public immediately (either anonymously or not) or keep private until they are ready to release it.

Pre-registration has seen widespread adoption in fields like experimental psychology, medicine, and economics over the last decade.
In this setting, it has several benefits.
First and most directly, pre-registration allows the reader of a scientific study to see what parts of an analysis were planned in advance versus decided upon later.
While this does not necessarily eliminate the type of data-dependent decision making that can lead to unreliable results, it indirectly discourages the practice.
Second, the ability to compare the contents of a paper to its corresponding pre-registration also provides readers with
a means of learning about possible null results that might otherwise not be reported due to publication biases.
Third, while not guaranteed by the process,  it has the ancillary benefit of encouraging analysts to think carefully about analyses before committing to them.
Anecdotally, many authors report that they see pre-registration as having intrinsic value for increasing the intentionality of their research process, independent of its other uses. 

We believe that pre-registration is a promising approach for predictive modeling, where issues like data-dependent decision making and test set leakage can render performance estimates invalid, but that it requires some specific modifications to address the concerns listed above.
To this end, we have developed a two-phase pre-registration protocol for predictive modeling, shown in Table~\ref{tab:empty-prereg-form}.

\begin{table}
    \centering

    \scalebox{0.95}{
    \begin{tabular}{|>{\raggedright\arraybackslash}p{0.17\textwidth}|>{\raggedright\arraybackslash}p{0.36\textwidth}|>{\raggedright\arraybackslash}p{0.42\textwidth}|} \hline 
        \multicolumn{3}{|p{\textwidth}|}{\vspace{3ex}\centering\textbf{Pre-registration (A): Problem definition}\vspace{3ex}} \\ \hline
        \textbf{\formRQ~Research question} & What problem are you studying? & Predicting who will survive the Titanic disaster. \\ \hline 
        \textbf{\formDV~Dependent variable} & What is the main outcome of interest and how is it measured? & Binary label (survived or deceased) in ground truth data. \\ \hline 
        \textbf{\formIV~Independent variable} & What features will you use to predict this outcome? & I'll use all the features available in my dataset (11 features). \newline I will perform feature engineering using recursive feature elimination (feature\_selection.RFE in scikit.learn). \\ \hline 
        \textbf{\formTrainData~Training data} & How was the training data constructed? & Historical information about passengers on the ship. \newline I will split this data to reserve a validate set equal in size to the test set. \\ \hline 
        \textbf{\formTestData~Test data} & How will the test dataset be constructed? Will new data be collected, or will the test set be synthetically constructed? & Randomly set aside 20\% of observations for test set. \\ \hline 
        \textbf{\formFiltering~Data transformations} & Will you filter or transform the data from its raw form in any way? & I'll remove the cases with missing values in any of the variables. \\ \hline 
        \textbf{\formMetrics~Metrics} & How will you measure the success of your model in predicting the outcome? If you are using a measure such as accuracy that requires thresholding of a continuous prediction, please specify the threshold(s) you will use. & Precision in the top 100 most highly predicted cases (classification, unthresholded) OR AUC (classification), precision w/ threshold of 50\% (classification). \\ \hline 
        \textbf{\formBaselines~Baselines} & Will you compare your method to other baselines? If so, which ones? & Logistic regression with the same features \\  \hline
        \textbf{\formOtherA~Anything else} & Is there anything else you would like to pre-register before training and validation? & No \\ \hline 
        \multicolumn{3}{|p{\textwidth}|}{\vspace{3ex}\centering\textbf{Pre-registration (B): Model details}\vspace{3ex}} \\ \hline 
        \textbf{\formPredMethod~Prediction method} & What prediction method(s) are you using? & Random forest classifier. \\ \hline 
        \textbf{\formModelTrain~Model training} & How did you train the model(s) you’re using? Please specify details such as cross-validation, resulting random seeds used, resulting hyperparameters, computational budget, etc. & Seed = 42, 100 trees, … (e.g., what are params for Sklearn's RandomForestClassifier). \newline 10-fold cross-validation (k=10). \\ \hline
        \textbf{\formTestAccess~Accessed test data} & Have you in any way previously accessed the test data? & No \\ \hline
        \textbf{\formOther~Anything else} & Are there any secondary or exploratory analyses you’d like to pre-register? & Look at accuracy broken down by gender/class/etc. \newline Investigate importance weights of features. \\ \hline
        \textbf{\formAdjustments~Plan adjustments} & Are there any changes in your process as compared to your previous answers that you’d like to report? & I dropped two additional features (sex, age). \\ \hline
    \end{tabular}
    }

    \caption{Questions for each of the two pre-registration phases, with example answers for the problem of predicting survivors of the Titanic (rightmost column). Phase (A), completed prior to any model training or validation, focuses on defining the problem, including the relevant variables and evaluation criteria. Phase (B), completed prior to model testing, captures details of the finalized model.
    }
    \label{tab:empty-prereg-form}
\end{table}

The first phase (A) is designed to be completed \textit{before} any model training or validation is done, with the purpose of capturing details of the researcher's goals for the modeling exercise, the high-level problem definition, and how it will be operationalized.
This includes a statement of the primary research question, along with declaring the dependent and independent variables, how the training and test sets will be constructed, whether any filtering or transformations are planned, what metrics will be used, and what baselines the model will be compared to. To be clear, researchers will almost always have to conduct \textit{some} exploratory analysis prior to phase (A)---for example, to familiarize themselves with the structure of the data, the presence of missing or incorrectly coded values, or the ranges of individual features. However, we draw a distinction between exploration of the properties of the data set itself and analysis that involves modeling relations between features and outcomes (e.g. by computing correlations). Whereas questions about the former can be answered prior to phase (A), questions about the latter should be answered only after phase (A) has been completed.
The reason for having the researcher answer model-related questions prior to training is to discourage adjustment of the problem definition based on the results of the modeling exercise.
Although, as described below, we allow researchers to update their answers to these questions after the training and validation phases, the hope is that encouraging researchers to be explicit about the high-level goals of a modeling exercise upfront---before seeing any modeling results---will reduce the gap between how a problem is framed versus operationalized.

After completing phase (A) of the pre-registration, the researcher conducts the usual training and validation steps from the traditional predictive modeling flow. This can include a simple train/validation split or K-fold cross-validation for model selection or hyperparameter tuning, and is often an iterative process.
When this process is complete and the model details are finalized, the researcher completes phase (B) of the pre-registration form \textit{before} testing the model.
This helps to cleanly separate model building from model evaluation and should reduce the prevalence of issues like algorithmic leakage and overfitting by multiple accesses of the test set.
Phase (B) includes declaring the specific prediction method to be used, including how the model was trained, any random seeds or hyperparameters used, and whether the test set has been accessed in any way.
This also allows the researcher to report any changes in responses to phase (A) of the pre-registration, with the understanding that the process of training and validation of a model often surfaces insights that lead to changes such as the addition or removal of features, or perhaps data filtering.
The objective of including this item in phase (B) is to avoid overly constraining researchers by allowing adjustments to the declarations in phase (A) if need be, but at the same time requiring that those adjustments are explicitly stated so that readers and reviewers can evaluate their potential impact accordingly.
Finally, following the format in the widely-used AsPredicted form for experimental work\footnote{\url{http://aspredicted.org}}, researchers can declare specific secondary or auxiliary analyses they have planned.

Building upon our outlined two-phase pre-registration protocol for predictive modeling, we pursued an initial investigation into how the protocol is interpreted and applied by practitioners.

\section{Observing the Protocol in Use}

As an initial investigation of the usability of the protocol, we conducted a qualitative study\footnote{Approved research study under IRB \#STU00219262 at Northwestern University, USA. Recruitment materials are provided in Appendix~\ref{app:appendixA}.} with six PhD students doing research in ML, whose expertise spanned general AI/ML (3), natural language processing (1), computer vision (1), and deep learning and nanophotonics (1). In these one-on-one sessions, we introduced participants to the protocol and then observed as they applied it to predict depression in teens according to a learning problem and dataset we provided. 
The purpose of these sessions was to put ML researchers in a situation where they had a real learning problem to solve, so as to better understand how natural (or not) they found it to apply the protocol, and whether it impacted their modeling choices.
We concluded each session with a semi-structured interview to assess perceived benefits and challenges. 

After summarizing the procedure below, we report on how participants used the protocol, including where they encountered challenges and where they perceived benefits.
While participants noted some challenges to widespread adoption of the protocol, all but one participant saw clear value in the procedure for improving the validity of ML-related research. 
Several participants commented on how the protocol aligned well with the general emphasis on obtaining valid test estimates in ML research.  
Moreover, we observed several direct instances where participants might have produced biased estimates of out-of-sample performance had they not adhered to the protocol, and in general observed that the ``forced reflection'' induced by pre-registration created opportunities for researchers to ground their modeling choices based on the research problem under consideration.

\subsection{Methods}

\noindent \textbf{Participants.} 
We recruited participants through the Northwestern Computer Science department's Ph.D. student listserv, requiring that participants had experience doing predictive modeling on a regular basis and %
were familiar with predictive modeling in Python.
Participants were scheduled for 90-minute Zoom sessions conducted by two of the authors, for which they later received a \$90 Visa gift card as compensation. %

\vspace{3mm}
\noindent \textbf{Learning problem.}
Participants were tasked with the problem of using responses from 2016 Monitoring the Future survey of 12th graders to predict depression in teens.
We provided participants with a subsampled version of the 2016 data adapted from that used by \citep{orben2019association}, which contained values of 12 potential outcome variables representing attitudes elicited via Likert-style items (e.g., ``Life often seems meaningless.''; ``I feel I am a person of worth, on an equal plane with others.''), and 24 potential predictor variables summarizing behaviors elicited via Likert-style items (e.g., ``How much TV do you estimate you watch on an average WEEKDAY?''; ``How often do you get at least seven hours of sleep?'').
Participants were told they would train a model and evaluate performance metrics on a held-out test set.

\vspace{3mm}
\noindent \textbf{Procedure.}
Prior to the session, we provided the participants with a description of the learning problem and all variables, as well as a version of the training dataset where the outcome variable columns had been shuffled. We provided shuffled training data so that participants, who we presumed were not familiar with the nature of the dataset, could examine the distributions of predictors and distributions of outcomes, but not their correlation prior to the session. 
Our goals were to encourage them to become familiar with the data prior to modeling while preventing the problem of simply finding a few highly correlated variables. Most participants (four) came to the session with some familiarity with the data variables and an idea of the type of model they would attempt.

\begin{figure}[htp]
    \centering\includegraphics[width=1.0\textwidth]{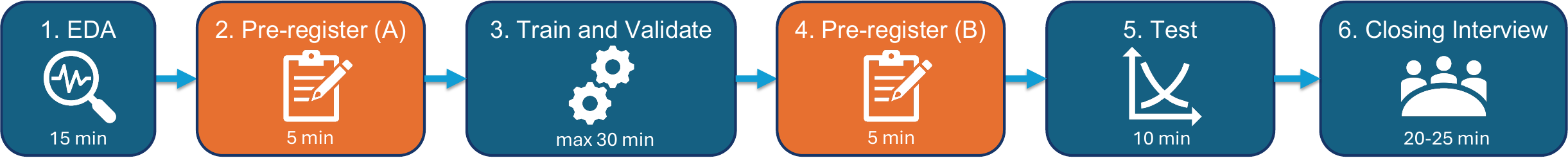}
    \caption{After a brief introduction, each session was divided into six phases with pre-defined time frames.}
    \label{fig:process}
\end{figure}

Each session started with the participant verbally consenting, followed by a brief description of the goals of pre-registration. We introduced the protocol, including example answers to each question for the problem of predicting survivors of the Titanic disaster.
We informed participants that the protocol sections needed to be completed before they began training or testing their ML models, respectively, and that they could not later go back and change their responses.
We next provided participants with a Google Colab notebook with designated cells for phases of the session (see Appendix~\ref{app:appendixB}), including visualization plotting code to facilitate initial exploratory analysis of individual variables in Appendix~\ref{app:appendixC}. The remainder of the session, along with the allocated time per phase, can be seen in Figure~\ref{fig:process}.

We kept participants informed of their remaining time per phase, and that they should expect to satisfice compared to their usual practice due to the limited time. We moved them to the next phase when time ran out. All six participants were able to finish training a model in the time allotted, though several participants skipped validation. Because our goal was to observe and collect feedback on the protocol, we emphasized that the performance they achieved was not crucial to our evaluation. We encouraged participants to share any questions they had and to verbalize their thought process as they worked, emphasizing that ``There are no right or wrong answers, so share all thoughts and concerns.''

\vspace{3mm}
\noindent \textbf{Interview questions.}
\label{sec:interview}
The concluding 20--25 minutes of each interview asked them to think about other recent projects they have done and imagine if they had been required to use the protocol. We then posed a few questions intended to get at their usual process for concluding training and validation, and how they tended to react to lower-than-expected test accuracy. 

Specifically, we asked: 
\begin{enumerate}
    \item Imagine you had more time and could edit your model or training process more during the training/validation phase. When would you stop editing and go forward to the test phase?
    \item Now, imagine you see the test accuracy, and it is lower than you expected based on validation accuracy. What would you do next?
    \item To what extent do you think data scientists like yourself would be interested in this pre-registration protocol?
    \item What do you see as the biggest obstructions to data scientists adopting this protocol? 
\end{enumerate}

While the 90-minute limit for the study constrained the amount of time participants had to learn about the data and develop a predictive model, the synchronous interview format gave us the ability to collect detailed, real-time feedback on participants' experience while they used the protocol, which would have been difficult to achieve through other means (e.g., assigning it to students as homework, or running an asynchronous online study).

\subsection{Results}

In reporting results, we summarize how participants approached the learning problem with a focus on how the protocol appeared to affect their actions in each phase of the study. We discuss the likely effects of the protocol based on participants' stated plans after they observed low test accuracy, and synthesize themes in their perceptions of the benefits and challenges of widespread adoption of the protocol.

\subsubsection{Exploratory data analysis}
All participants used the provided visualization code to assess distributions for missingness and explore the data to inform feature selection.
At the beginning of the EDA phase, we encouraged participants to select a single dependent variable to simplify the problem under the time constraints. All participants observed that most of the potential dependent variables had a high proportion of missing values. Out of six participants, two selected the variable with considerably fewer missing observations (Taking all things together, how would you say things are these days---would you say you’re very happy, pretty happy, or not too happy these days?) while four chose other dependent variables. \textit{P4}, \textit{P5}, and \textit{P6} debated whether to binarize the problem to balance the classification or to predict individual classes; ultimately, all but one participant committed to treating the problem as classification rather than regression. Given that we had to remind multiple participants to commit to their approach prior to proceeding to model training, it seems likely that without pre-specification, several would have tried both classification and regression and used training performance to decide.
More details can be found in Appendix~\ref{app:appendixD}, which highlights the variety of different ways that participants went about the modeling exercise.

\subsubsection{Phase A: Before training} \label{sec:beftrain}
While most participants seemed to find it easy to fill out the first protocol, a few had trouble making decisions and/or overlooked that their answers were meant to specify an exact plan rather than a gesture toward options they would try. The research question (A.1) and test set (A.5) were provided. The training data (A.4) was also pre-defined: while we reminded all participants as they filled out the form that they would need to declare further separation of training data if they wanted a separate validation set, all participants but two used k-fold cross validation, which does not require an additional split, and one participant did not validate at all. This left questions about independent (A.3) and dependent variables (A.2), data transformations (A.6), and metrics (A.7).

When it came to specifying the dependent variable (A.2), \textit{P3} was unsure whether to treat the problem as a classification or regression task and wanted to pre-register both options, though ultimately opted for classification out of simplicity. 
Had \textit{P3} decided to pre-register both options, it would be important that they specify how they planned to report the results, e.g., reporting both models or selecting which to prioritize using some a predetermined decision rule.

Additionally, several participants needed to be reminded to commit to their data variable choices (A.2, A.3) and/or data transformation choices (A.6) prior to proceeding to the training phase.
\textit{P5} expressed interest in using automatic feature selection processes to filter and engineer features, while \textit{P4} thought of using principal component analysis (PCA) for feature selection but without detailing the algorithmic specifics. 
\textit{P2}, \textit{P4}, and \textit{P5} struggled in specifying an imputation strategy. For instance, \textit{P2} considered pre-registering both the mean and mode. 
The act of filling out the protocol led such participants to reflect on their intentions in ways that they seem unlikely to have done otherwise.

In specifying performance metrics (A.7), all participants noted intentions to use standard classification or regression metrics (e.g., precision, recall, F1). However, some overlooked the need to specify whether they intended to report metrics at the level of classes (when the outcome had more than two classes) or over classes (e.g., weighted average). %
Additionally, one participant who opted to binarize a Likert-style question to create the outcome needed to be reminded to specify how exactly they would perform the dichotomization.

We omitted pre-specification of Baselines (A.8) from Phase A in the study for the purposes of these interviews, as we did not presume they had exposure to any existing literature on this problem or knowledge of state-of-the-art methods for it.

\subsubsection{Model training and selection} 
Because the protocol does not require committing to a particular model class, participants were free to specify whatever type of model they wished. Due to the limited time however, most participants did not explore more than one type of model, instead opting for what they deemed most promising given the learning problem and their knowledge of the data. Most participants used a form of model ensembling. While many participants used k-fold cross-validation to mitigate overfitting, there was some variation, with \textit{P1} skipping validation entirely.

A few participants realized they had overlooked complications in filling out Phase A once they reached the modeling phase. For example, \textit{P2} was thinking of normalizing all features to a single scale of 1--5 to have the same range of values, but they did not commit to a strategy early on. During training, they
decided to normalize the selected features to a 0 to 1 range. Another example is \textit{P4}, who realized that classes were imbalanced when looking at the confusion matrix; when they sampled with weights, the validation accuracy decreased. Upon seeing these results, they concluded that binarizing the outcome variable would have made the imbalance easier to address.
Similarly, \textit{P4} was uncertain about which strategy to adopt for transforming features. They began by considering simple imputations, such as taking the mean, but then considered more complex processes, like handling data based on their explanatory power (e.g., dropping a variable if it contributes less than 1\% of the variance to the dataset). %
Though they had pre-registered using the mean, they later decided that using the mode was a better approach but stuck to their initial plan due to the protocol. %

\subsubsection{Phase B: After validation, before testing} \label{sec:beftest}

Completing Phase B of the protocol was quicker and more straightforward than Phase A for most participants. All participants described their prediction method without trouble (B.1). In describing model details (B.2), participants mostly felt it was clear what they should report, although our use of the Google Colab environment reduced the thought they had to put into specifying the computational budget.
When \textit{P5} got to Phase B, they realized they had overlooked setting a random seed, and wanted to go back, change it, and then register this modification. 
However, there was no need to specify anything differently, as they could simply go back to model training and return to Phase B after retraining with a seed.
\textit{P6} used the `Plan adjustments' field (B.5) to report on their use of the `micro' average function parameter because they did not remember to specify the specific method in Phase A.

Because participants were only given access to the test data when they reached the test phase of the session, all could describe the test data as previously unaccessed (B.3), though as we later describe, some participants commented in the interview that this might sometimes be a difficult question to answer in practice.

Recall that Phase B also allowed participants to pre-register any secondary analyses (B.4) before accessing the test set, as is included in conventional (non-ML) pre-registration. Two participants expressed an interest in describing secondary analyses. \textit{P5} added the requirement of checking feature importance after testing with the idea that doing so would inform their next steps. %
\textit{P4} mentioned that time allowing, they would use B.4 to pre-register an analysis of how their outcome variable correlates with other potential dependent variables and whether the same input features are as useful there by looking into weights in the SVM classifier.

\subsubsection{Interview results: Strategies regarding test set access} \label{sec:adjust}

When asked when they expected they would have felt comfortable progressing to the test phase had there not been time constraints (Section \ref{sec:interview}, Q1), 
multiple participants described how they would ideally have explored multiple models, and progressed when strategies like hyperparameter tuning no longer improved pre-test accuracy.

When asked how they would proceed if the test results were lower than expected based on pre-test estimates (Section \ref{sec:interview}, Q2), most participants described strategies aimed at identifying reasons for the poor performance, most commonly by comparing the test and train distributions to check if the test sample was deviant.
However, we observed a split when it came to how willing participants were to re-access test data in attempting to reconcile poor test performance: half of the participants made clear they would respect the pre-registered plan and not report performance statistics based on repeated test set use, while the other half described procedures that if not fully reported, might compromise the validity of the test estimates.

Of those who were hesitant to re-access test data, \textit{P3} reflected on possibly adjusting the training set to fix the class imbalance, but they expressed that the goal should be to investigate underlying issues in the data rather than altering the model and trying again. Similarly, while \textit{P5} expressed interest in evaluating feature importance for better predictive power, like \textit{P3}, they seemed to recognize the importance of staying truthful to the initial hypothesis if the estimates were to be valid.
\textit{P4} was the most adamant in acknowledging the threats of deviating from the pre-registered plan and described how negative results should be taken as a learning opportunity, not a push to attempt further model optimization. %

The other three participants described strategies that involved re-accessing test data. Such steps would need to be explicitly reported as data exploration in communicating the results of research; if not reported, they would jeopardize the reproducibility of the reported test metrics. \textit{P1} described how they might try reducing the model size, in terms of the number of neurons, in addition to checking the data distribution. If the model continued to fail after such checks would they attempt to acquire new data. %
\textit{P2} described how they would try varying the hyperparameter settings in cross-validation, and if this failed to improve test performance, systematically test the effects of feature combinations on performance, followed by rethinking their pre-processing steps.
\textit{P6} stressed the importance of cross-validation scores but also the value of examining the test data accuracy to understand if changes to the process helped. They said that they would first search for related work to select score thresholds based on SOTA performance benchmarks to provide a baseline for how much improvement was still reasonable. %
Both \textit{P1} and \textit{P6} believed it should be possible to continue updating and iterating on their pre-registration, which we explained would contradict the goals of pre-registering.
Hence, despite using the protocol, these three participants seemed to overlook the importance of only accessing the test data once.
The fact that two of these three participants did express some benefits of pre-registration (described below) suggests that helping some researchers to grasp the reasoning behind pre-registering may require more training than we provided in the restricted time of our sessions.

\subsubsection{Interview results: Attitudes on adopting pre-registration} \label{sec:chalopp}

\noindent \textbf{Perceived benefits.}
When asked to reflect on the potential benefits or challenges they expected if the pre-registration protocol were widely adopted, all but one of the participants described benefits.
Four participants specifically mentioned the value of reducing the equivalent of ``p-hacking'' in machine learning, including making decisions about outcome variables and metrics after seeing results (\textit{P2}, \textit{P4}, \textit{P5}, \textit{P6}).
When asked whether they believed the protocol would improve the validity of ML practice based on what they perceived as an ML researcher, \textit{P4} replied that it absolutely would.
A couple of participants alluded to the value of the protocol as a contract that would encourage researchers to be honest (\textit{P3}, \textit{P4}).
Several described how pre-registering would result in more thoughtful and/or structured approaches to learning problems (\textit{P2}, \textit{P3}, \textit{P4}) and greater transparency in research reporting (\textit{P4}). \textit{P2} commented that the approach was well aligned with the general philosophy in ML practice of having an initial idea, selecting baselines, and then holding oneself to the constraint of a held-out set.

\textit{P2} mused about how the protocol might affect problems caused by a non-stationary data-generating process, such as by encouraging deeper reflection on what aspects of a phenomenon are most predictable. 
Ultimately, though, \textit{P2} concluded that while pre-registration \textit{might} improve out-of-distribution performance, it was difficult to know whether, on average, this would be a benefit.

\vspace{3mm}
\noindent \textbf{Perceived drawbacks.}
All participants described at least one obstruction they saw to the widespread use of pre-registration.
The burden of time and energy required to adhere to the protocol was mentioned by several participants (\textit{P1}, \textit{P3}). \textit{P1} was concerned that pre-registering would limit creativity in modeling; although, as described above, \textit{P1} comments suggested they would have deviated from the protocol by re-accessing test data if the analysis continued, suggesting they did not fully grasp the concept of pre-registration. 
Both \textit{P1} and \textit{P6} expressed the need for an update mechanism to adjust a pre-registration later in the process to allow for flexibility, suggesting they had not fully internalized the motivation behind a time-stamped plan.

Other comments described specific ways in which pre-registering was perceived as at odds with typical practice. \textit{P6} described how it can be hard to identify the best metric to represent model performance a priori, suggesting that one could ``develop the model with well-accepted metrics for the problem but then later find a better metric that justifies how the model performs,'' but did not provide any specific examples of what they imagined.
\textit{P5} felt that pre-registering a feature engineering process in Phase A was difficult because their usual pipeline included a phase between initial perusing of the training data and model training in which feature processing is the sole focus.

Relatedly, \textit{P6} and \textit{P4} commented on the influence of prior knowledge in the process of pre-registering. \textit{P6} believed that when the analyst possessed little prior knowledge about the dataset, they would be more likely to make mistakes that might warrant deviating from their original plan. On the other hand, they noted, it is difficult to account for the potential influence of prior exposure to test instances, e.g., from having used the dataset for a different purpose or having been exposed to prior research that used it. 
In reflecting on the protocol as it would have applied to their prior research, \textit{P4}, who was quite enthusiastic about the value of pre-registration, seemed concerned that they could not have applied the protocol to a previous analysis that led to important insights about the behavior of transformer networks. However, \textit{P4}'s concern was resolved when they realized aloud that the prior work they had in mind was exploratory and thus not the target of pre-registration. %

Finally, several participants pointed to limitations in the ability of pre-registration to guarantee valid results.
\textit{P5} commented on how a motivated researcher could pre-register then deviate without others necessarily noticing that they had. 
\textit{P6} expressed several concerns about the ambiguity around whether a researcher who pre-registered had actually stuck to the specification, noting, for example, that a researcher could use different training or test data than they described and that especially when data is subject to privacy protections, it is difficult to know exactly how data were split.
\textit{P4}, on the other hand, acknowledged that it was the author's responsibility to honor their pre-registration as it provided no strict guarantees without intentional use.

\vspace{3mm}
\noindent \textbf{Potential changes to the protocol.}
Several participants suggested revisions to the protocol. %
\textit{P3} recommended moving the question about prediction method(s) (B.1) from Phase B to Phase A.
However, Phase A focuses on the definition of the learning problem. Unless the model class is an important aspect of the problem definition (e.g., the research question concerns how a particular model class will perform), it is unnecessary to pre-specify in order to obtain internally valid test estimates.

In response to improving the second form, \textit{P4} recommended noting any behaviors that emerge in the validation data for following up on in the test data, as a means of specifying patterns that were originally unexpected but could gain some validity if repeated in test data prior to it being accessed. This option aligns well with the reasoning behind P2.4, ``Are there any secondary or exploratory analyses you’d like to pre-register?''

\textit{P6} was concerned about the potential for pre-registration to lead to information leakage and copying of ideas, though this can be resolved in practice if users register the timestamped protocol online but control when it becomes public, as is enabled by existing pre-registration servers. 
Related to their emphasis on the risk of pre-registration being used as a signal of robust results without being properly adhered to, \textit{P6} proposed a more detailed verification process involving logging and timestamping of actions during data preparation and model training to ensure that the data used aligns with what the author pre-registered. %

\section{Discussion}
Our work suggests the potential for a lightweight, two-step pre-registration protocol to improve the reliability of results reported in predictive modeling.
The protocol we propose can be used similar to how pre-registration is implemented on existing pre-registration servers used by social scientists\footnote{See \url{https://aspredicted.org}, \url{https://osf.io/registries}.}, where each document is time-stamped prior to proceeding further in the research process. The completed forms therefore provide a record of the intended analysis that reviewers and others can compare against the completed analysis the authors report to diagnose potential sources of data-dependent decisions.
Our observations of ML graduate students' use of the protocol suggest that pre-registration could prevent overfitting due to data-dependent decisions and reduce test set re-use, provided it is adhered to, and that ML researchers may value the degree of intention it encourages.
We saw indications that pre-registration induced more critical thinking around the design choices participants made in addressing the research question at hand, and that it drew attention to the many different ways that a given question can be operationalized as a modeling task.

Based on our observations in the study and more generally, we believe that pre-declaring important aspects of a predictive modeling pipeline is not common practice in predictive modeling, but should be whenever the validity of performance estimates on held-out data is of interest. Impediments to pre-specification, including most of the participants' struggles in filling out the first part of the protocol, arise from the difficulty in pre-determining the best way to translate a high-level learning problem into a concrete specification. 
It is worth remembering that adopting pre-registration does not equate to eliminating all data-dependent decision-making, and that having to deviate from a pre-registration does not mean it was a waste of time~\citep{simmons2021pre}. 
For this reason, we intentionally built flexibility into  
the two-part protocol, where participants can report any deviations from the plan in phase (A) prior to testing.
Likewise, if there are deviations in phase (B), this does not invalidate the work, but simply means that researchers should clearly state those deviations when reporting their results so that readers can distinguish between planned versus executed analyses and make informed decisions about the contribution of the work.
This allows readers to distinguish between confirmatory and exploratory work.
It also needn't be the case that the research process ends after phase (B).
Inevitably, work that aims to confirm some hypotheses may generate new ones, which can and should be investigated with subsequent pre-registered studies.

The challenges of pre-specification are likely to be affected by the style of research contribution. When researchers are attempting to advance SOTA on a standardized learning task, such as defined by a benchmark, degrees of freedom are likely to be reduced relative to when a researcher is studying, for example, the predictability of some social phenomenon.   
Indeed, such standardization has been deemed a unique contributor to success in empirical ML~\citep{donoho2023data}. 
Likewise, the benefits of pre-registration depend on the scenario as well.
In particular, we expect that the process outlined here will be most useful in cases where it is difficult or impossible to obtain truly new, out-of-sample data, as these settings are vulnerable to problems such as leakage and multiple test set accesses.
In contrast, for settings where models are deployed in production and new data are continually being collected (e.g., from one day to the next), it is likely that overfitting to the test set will eventually be uncovered by practitioners.

Additionally, where degrees of freedom in translating from goals to implementation are unavoidable, researchers may encounter fewer challenges in adopting pre-registration over time, as they acquire more experience using it. 
Our own experiences adopting pre-registration for exploratory modeling suggest that 1) getting in the habit naturally shifts one's practice to prioritize advance pre-specification of critical components of an analysis, and 2) with increasing familiarity one develops a better sense of what is useful to pre-specify, and at what level of detail. 
For example, some participants wondered if they should pre-specify the model class they intended to use, but this information is not necessary to ensure the validity of the performance estimates. 
Similarly, struggles around how to determine feature engineering in advance might become easier with experience, as the researcher learns when it is best to pre-specify a process versus the outcomes of that process (e.g., specific variables).

There is likely to be value in greater integration of reform philosophies and empirical evidence around reproducibility, replication, and generalizability, including pre-registration, in graduate curricula fields that practice predictive modeling.
As techniques from causal inference and social science are increasingly brought to bear on predictive modeling problems, 
it becomes important to translate what is known about causes of irreplicability and irreproducibility from social science to ML-based science~\citep{hullman2022worst}, so that those interested in integrative modeling~\citep{hofman2021integrating} know what to expect.
For example, evidence of the consequences of tailoring an analysis specification to achieve desired results and undisclosed degrees of freedom on the literature in fields provided through large-scale replication attempts (e.g.,~\citep{camerer2018evaluating,nosek2015}) can prompt valuable reflection on similar threats in predictive modeling despite ongoing discussion of how to best define and interpret replication failures~\citep{buzbas2023logical}. 
Anecdotally, we have found in our own classes at Northwestern, Columbia, and Penn that students working with predictive modeling are interested in being exposed to research on concerns with replicability from other fields, and that such exposure can lead to promising new lines of research.

Having made the case for the importance of pre-registration in predictive modeling, this leads to the question of how best to implement pre-registration such that it is easily adopted by those doing research, and accessible and verifiable by those consuming research.
Again, there is likely much to learn from various pre-registration frameworks that exist in the explanatory modeling community, such as the Open Science Framework (OSF) and AsPredicted.org.
At the same time, it may be the case that there are innovations specific to the predictive modeling world that could offer alternative implementations.
For instance, given that it is common for predictive modeling projects to utilize version control while developing software to build and test models, one could imagine that simply committing timestamped versions of the pre-registration forms for phases (A) and (B) to a code repository could serve a very lightweight but effective procedure.
Similarly, given that there are several existing frameworks for reporting details of predictive models, one could imagine automatically generating or linking Datasheets~\cite{gebru2021datasheets} and/or Model Cards~\citep{mitchell2019model} to pre-registrations for a given project.

\subsection{Limitations of Pre-registration}
It is important to recognize that pre-registration is not a panacea, and that just because a study incorporates pre-registration this does not necessarily make it a rigorous or important study.
Like many proposed reforms to increase the validity of scientific estimates, pre-registration depends on proper application, including buy-in by both authors and audiences.
For pre-registration to serve its intended purpose of improving replicability, authors must report all analyses from a pre-registration along with any deviations they make from it.
Pre-registration may not lead to obvious improvements if the  authors who are likely to pre-register are also a priori more likely to be intentional and transparent in their research, and that those who could benefit the most will fail to take advantage of pre-registration. 
Likewise, readers and reviewers must take the time to critically compare a study to what is planned in its accompanying pre-registration, requiring the development of new norms for consuming research.

Furthermore, while it is our hope that pre-registration for predictive modeling will curb data-dependent decision making and test set leakage, there are many other aspects to designing a good study that pre-registration does \textit{not} directly address.
For instance, one can pre-register a predictive model that optimizes a metric which is a poor proxy for success in a real-world deployment, or evaluates performance on an unrepresentative data set.
Such concerns are reminscent of critiques of pre-registration in explanatory modeling as often irrelevant to larger problems of weak theory~\citep{szollosi2019preregistration,szollosi2021arrested}.
There are also broader issues about generalizability~\citep{johnson2018generalizability,recht2018cifar,recht2019imagenet,singh2022generalizability,yarkoni2022generalizability} that pre-registration does not necessarily address.
One reason that even pre-registered results may fail to generalize is due to distribution shift in covariates, targets, or the relationship between them~\citep{quinonero2008dataset}.
Whenever the training and test data are drawn from different distributions, making reliable statements about test set performance can be challenging.
Such failures are difficult to detect in studies that utilize synthetic train/validate/test splits of pre-defined datasets, but are commonly encountered when deploying models, as the (social) world is highly contextual and non-stationary~\citep{lazer2014parable}.
This is coupled to issues about the robustness of findings, where loose verbal assertions imply that methods will produce similar results under new conditions when this is not actually the case.
For example, too easily equating good performance on a benchmark with human-like mastery of a task (e.g., ``reading comprehension,'' ``object recognition'') threatens the validity of claims about performance outside of the specific dataset used to compute the performance statistics~\citep{liao2021we,lipton2019research,recht2019imagenet}.

Separately, there is the issue of publication bias, or the ``file drawer problem''~\citep{rosenthal1979file}, where negative or null results may not be shared or published as widely as positive ones.
It is, of course, possible that many pre-registered studies might never be published.
Registered reports, which incorporate pre-registration into the peer review process, are a proposed solution to this issue~\citep{chambers2013registered,nosek2014registered}.

It is important that both authors and readers recognize these limitations, so that pre-registration does not become a potentially unreliable signal for scientific rigor or motivate researchers or readers to think less carefully about other aspects of a study.
Unthinking reliance on heuristics is both a cause of problematic scientific studies and a risk of proposing new solutions.
On the contrary, pre-registration should encourage more (rather than less) critical thinking when designing and evaluating studies.

We think that pre-registration for predictive modeling is best seen as a tool for stimulating greater foresight and helping authors to distinguish between planned and unplanned analyses, rather than an ironclad solution to broad issues in the field. 
The challenges pre-registration may help with are diverse: 
issues of cherry-picking comparisons to support a foregone conclusion, for example, are different from issues of leakage. 
More broadly, the distinction between exploratory and confirmatory research is often blurry for good reason~\citep{szollosi2019preregistration}. 
Blanket arguments that data re-use is always bad misrepresent the nuance that often characterizes statistical modeling, where the answer about whether knowledge of data is harmful to inference is often ``it depends''~\citep{devezer2020case}. 
That said, valid test performance statistics are central to many applications of predictive modeling (e.g., reducing leakage, multiple accesses to test sets, and over-tuning of model hyperparameters), such that some of the potential benefits of 
pre-registration are hard to dispute.

\subsection{Related Work}
Our work contributes to a broader landscape of reform procedures that have been proposed to improve the reproducibility of research that relies on predictive modeling. 
Perhaps most common have been proposed reporting guidelines and consensus-based checklists for authors to follow, including general ML reporting guidelines for data~\citep{gebru2021datasheets,rogers2021just} and models~\citep{mitchell2019model}, ML reproducibilty checklists~\citep{gundersen2018reproducible,pineau2021improving}, and other reporting advice for eliminating current confounds, such as reporting validation performance as a function of computation budget~\citep{dodge2019show}.
Other checklists are tailored to certain domains, such as CLAIMS for medical imaging-based research~\citep{mongan2020checklist} and REFORMS~\citep{kapoor2023reforms} for research that uses ML model performance to make scientific claims, excluding ML methods research. These methods aim to improve reproducibility through more transparent reporting. In contrast to pre-registration, these approaches do not necessarily aim to change how researchers plan or conduct their research.

Researchers have also proposed specific changes to evaluation procedures, such as using random train-test splits across experiments~\citep{gorman2019we}, doing proper power analysis~\citep{card2020little}, and conducting a coordinated evaluation of algorithms for reinforcement learning~\citep{khetarpal2018re}, to name a few. Such procedures are intended to produce more robust evaluation results, but address methodological weaknesses that are orthogonal to the types of data-dependent decisions that pre-registration aims to reduce.

Closer to our goals, pre-registration workshops were hosted at NeurIPS 2020~\citep{bengio202opre} and 2021 Pre-registration workshops~\citep{albanie2021pre}.
However, these workshops defined pre-registration as ``reviewing and accepting a paper before experiments are conducted,'' thus interpreting the concept as what has more widely been termed a registered report~\citep{chambers2013registered, chambers2015registered,chambers2022past}. In a registered report, authors propose an idea but wait until the proposal is designated as involving a worthwhile research question via peer review to run the experiments and report the results. Registered reports are thus harder to implement in practice, as teams of qualified reviewers must be found. In contrast, pre-registration operates independently of the peer review system, allowing authors to time-stamp a record of the planned analysis prior to the conduct of the research, which they can later make public.
So long as a third party is willing to time-stamp a document, pre-registration is feasible regardless of available reviewer pools.

\bibliography{ref}

\newpage
\begin{appendices}
\section{}
\label{app:appendixA}

\begin{center}

\includegraphics[angle=-90,origin=c,width=0.7\textwidth]{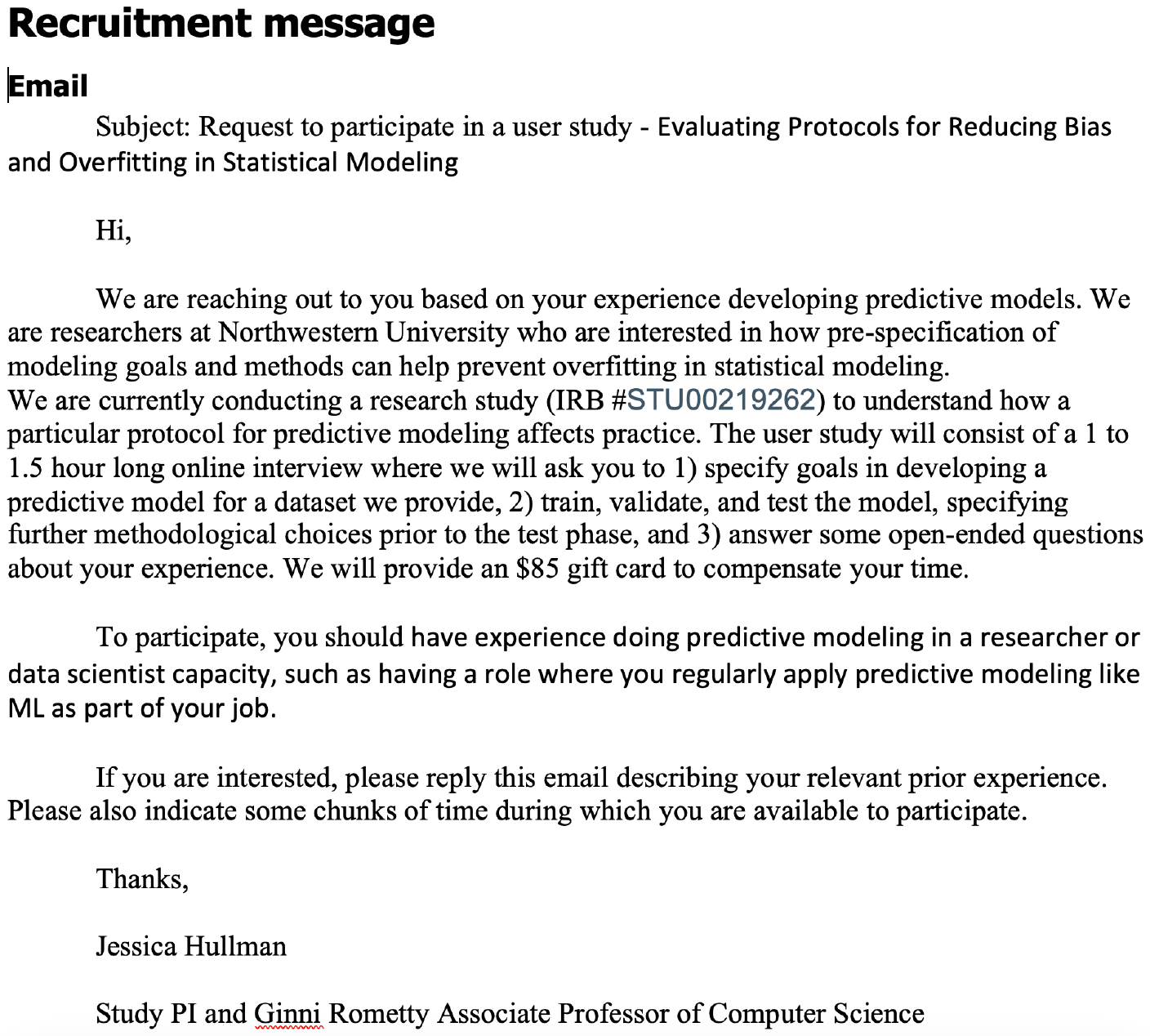}
\clearpage
\includegraphics[page=1, scale=0.8]{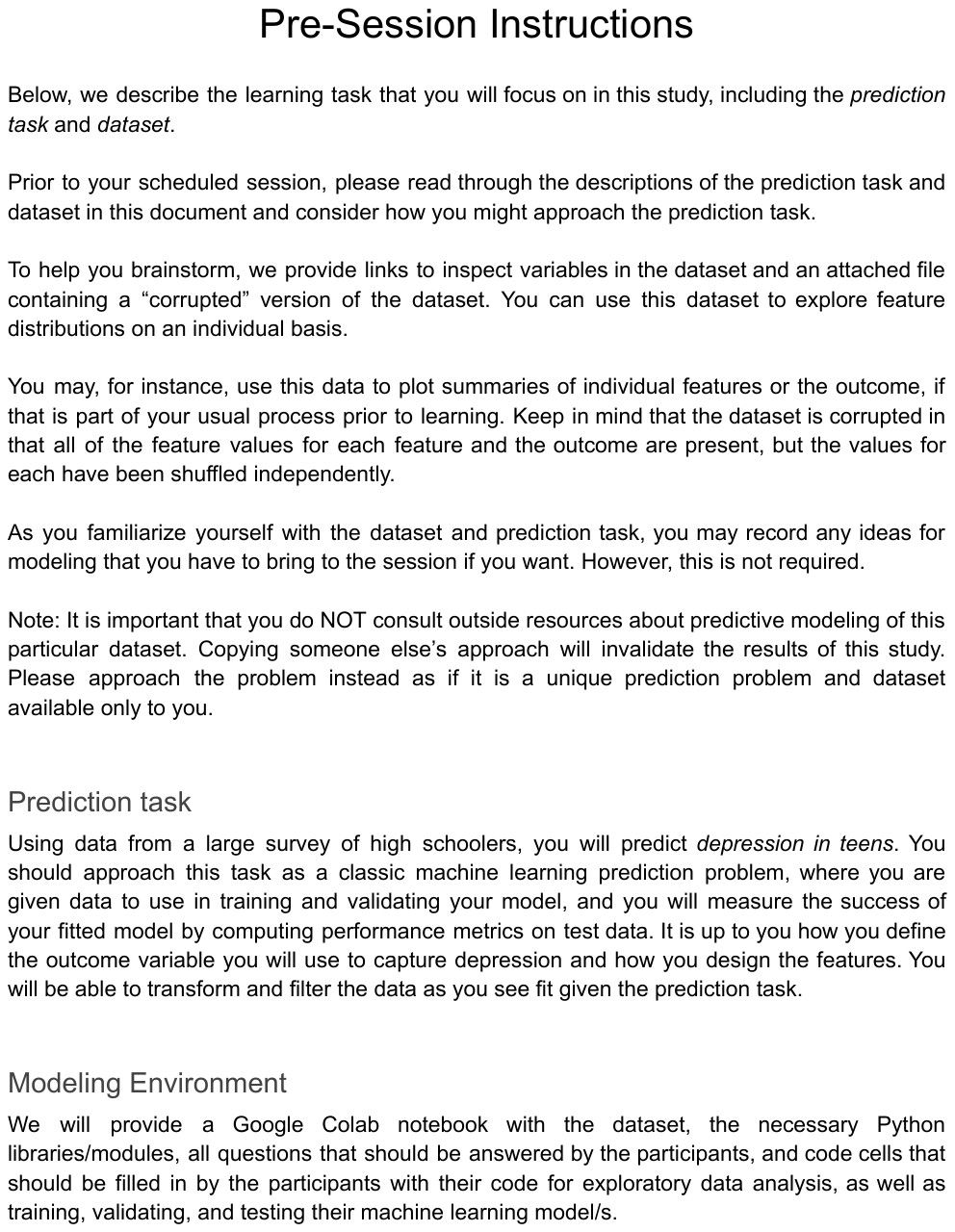}
\includegraphics[page=2, scale=0.8]{figures/prel_instr.pdf}    
\includegraphics[page=3, scale=0.8]{figures/prel_instr.pdf}  
\includegraphics[page=4, scale=0.8]{figures/prel_instr.pdf}
\includegraphics[page=5, scale=0.8]{figures/prel_instr.pdf}
\includegraphics[page=6, scale=0.8]{figures/prel_instr.pdf}
\end{center}

\setcounter{figure}{0}
\counterwithin{figure}{section}
\section{}
\label{app:appendixB}

\begin{figure}[htp]
    \centering\includegraphics[width=0.7\textwidth]{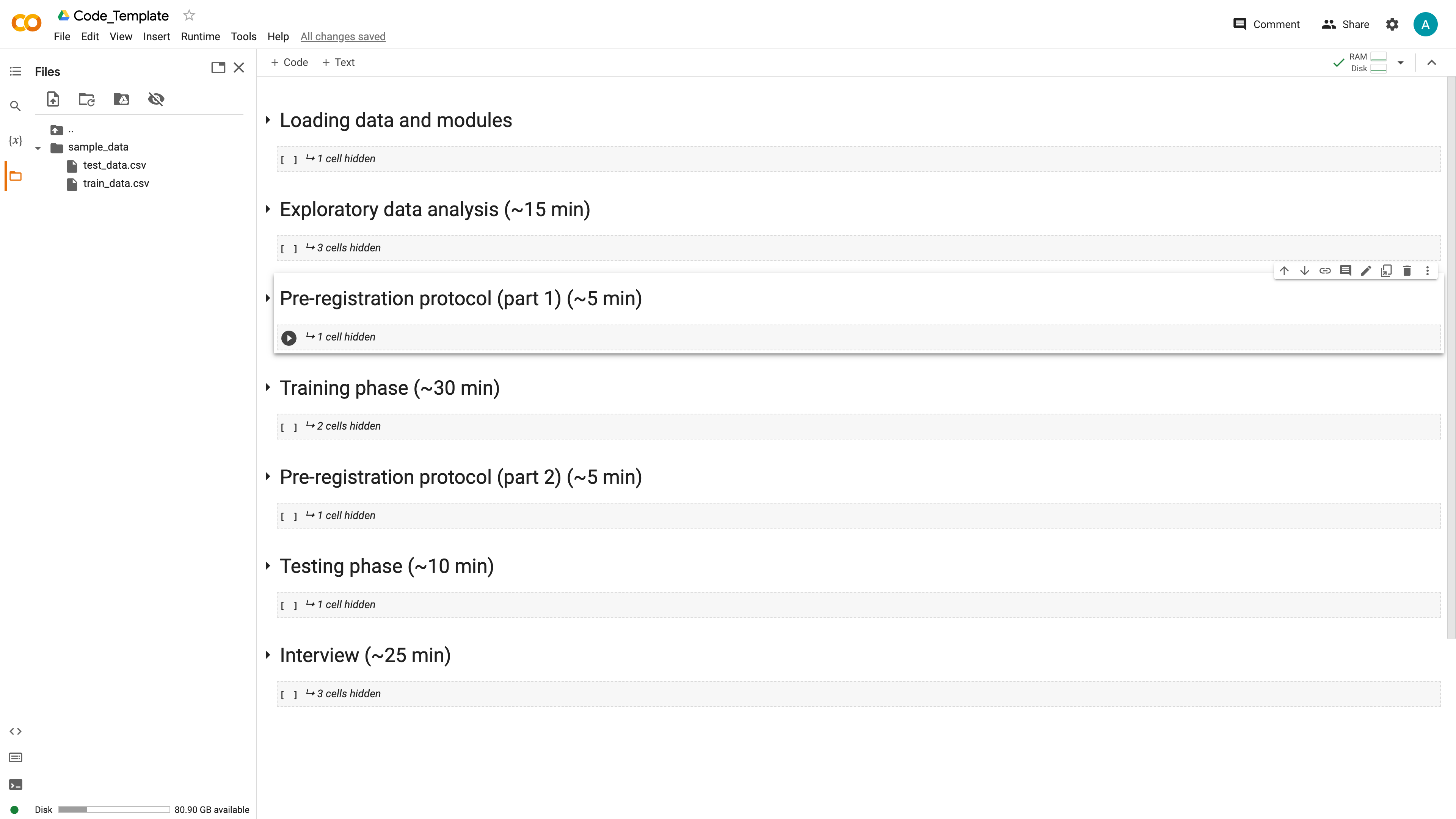}
    \caption{The Google Colab code template given to each participant during the main experimental phase.}\vspace*{-5pt}
\end{figure}

\setcounter{figure}{0}
\counterwithin{figure}{section}
\section{}
\label{app:appendixC}

\begin{figure}[htp]
    \centering\includegraphics[width=0.7\textwidth]{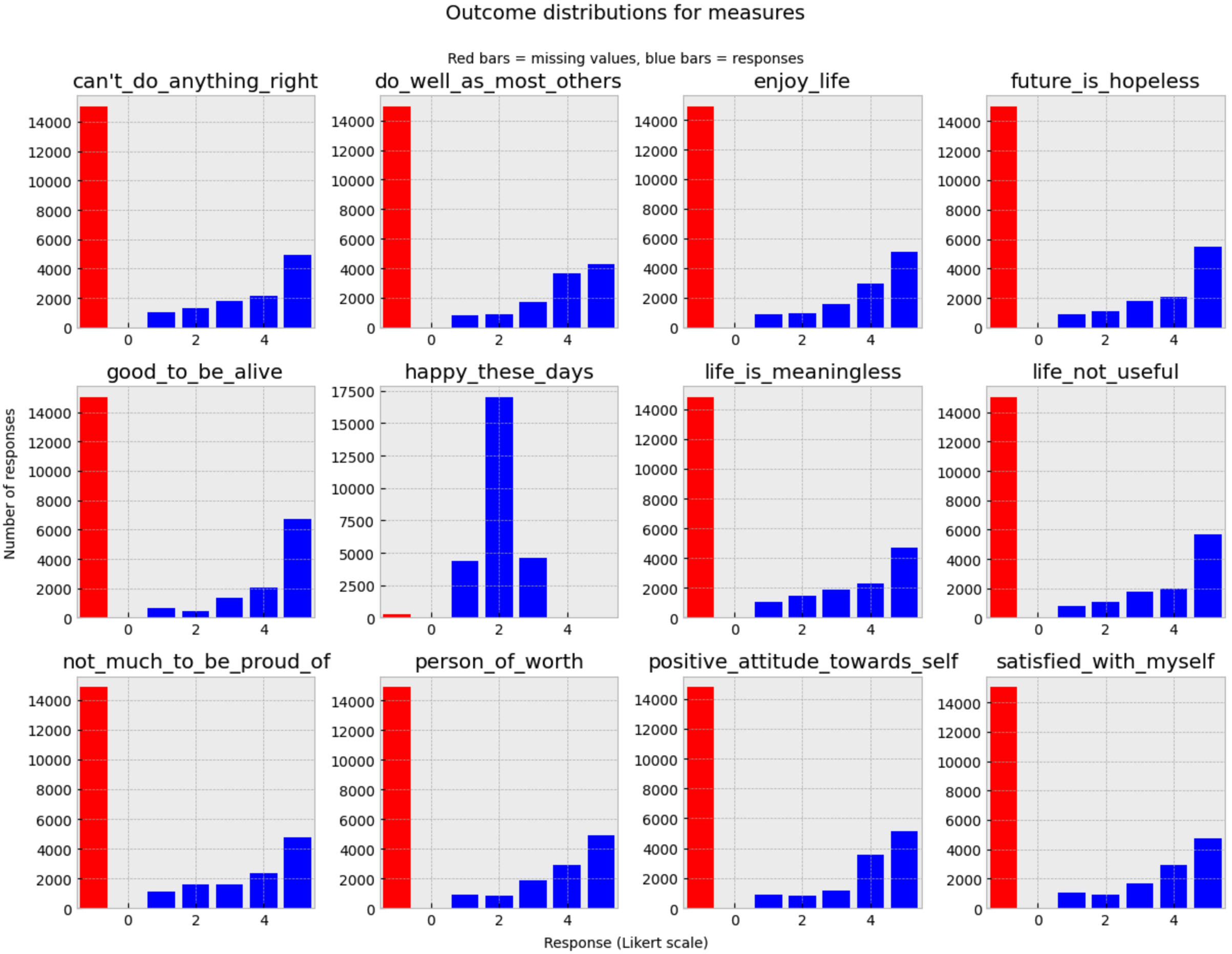}
    \caption{Missing values (\textcolor{red}{red}) and distributions (\textcolor{blue}{blue}) of the 12 different dependent variables.}\vspace*{-5pt}
\end{figure}

\begin{figure}[htp]
    \centering\includegraphics[width=0.86\textwidth]{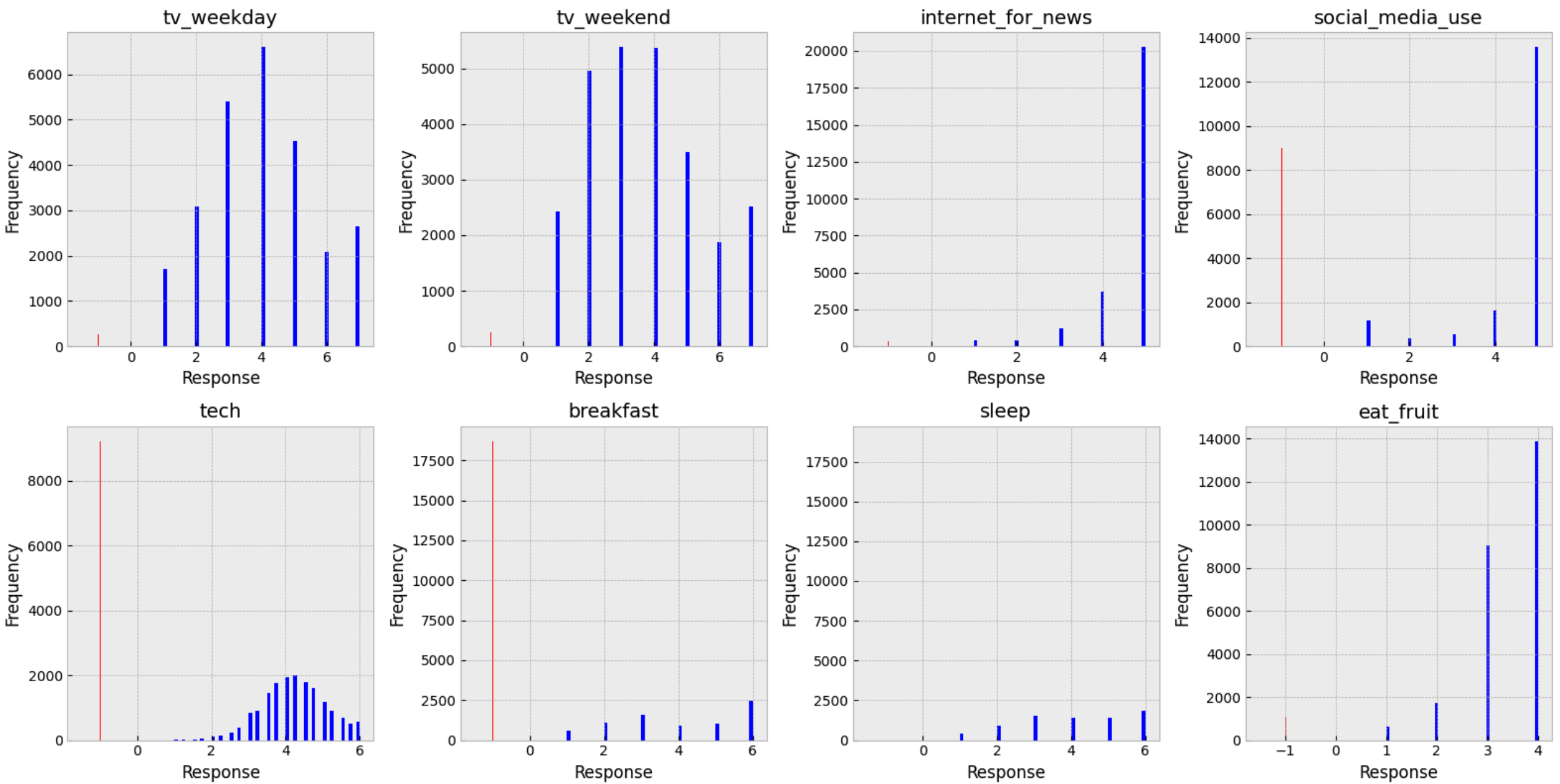}
\end{figure}

\begin{figure}[htp]
    \centering\includegraphics[width=0.86\textwidth]{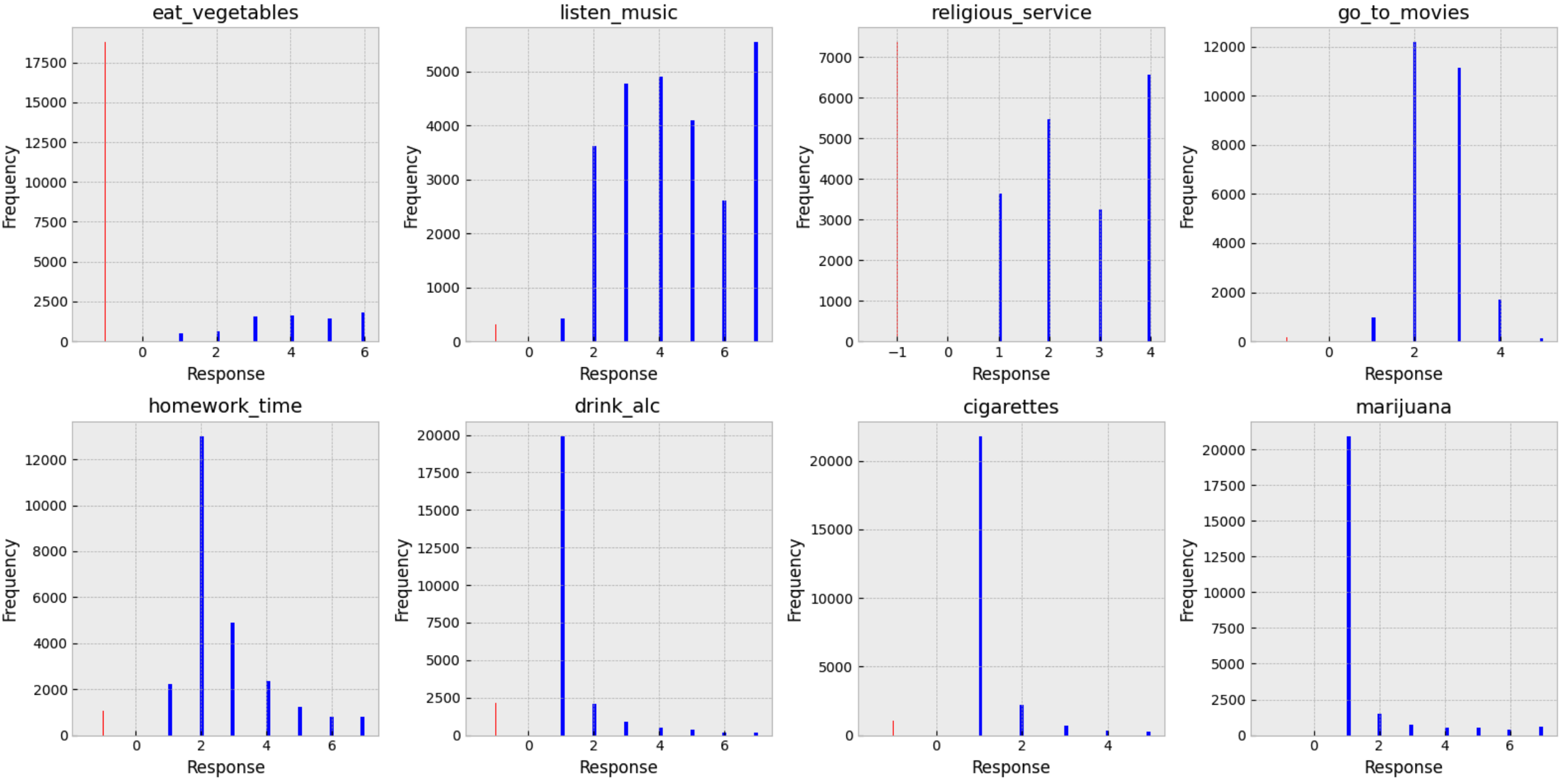}
\end{figure}

\begin{figure}[htp]
    \centering\includegraphics[width=0.86\textwidth]{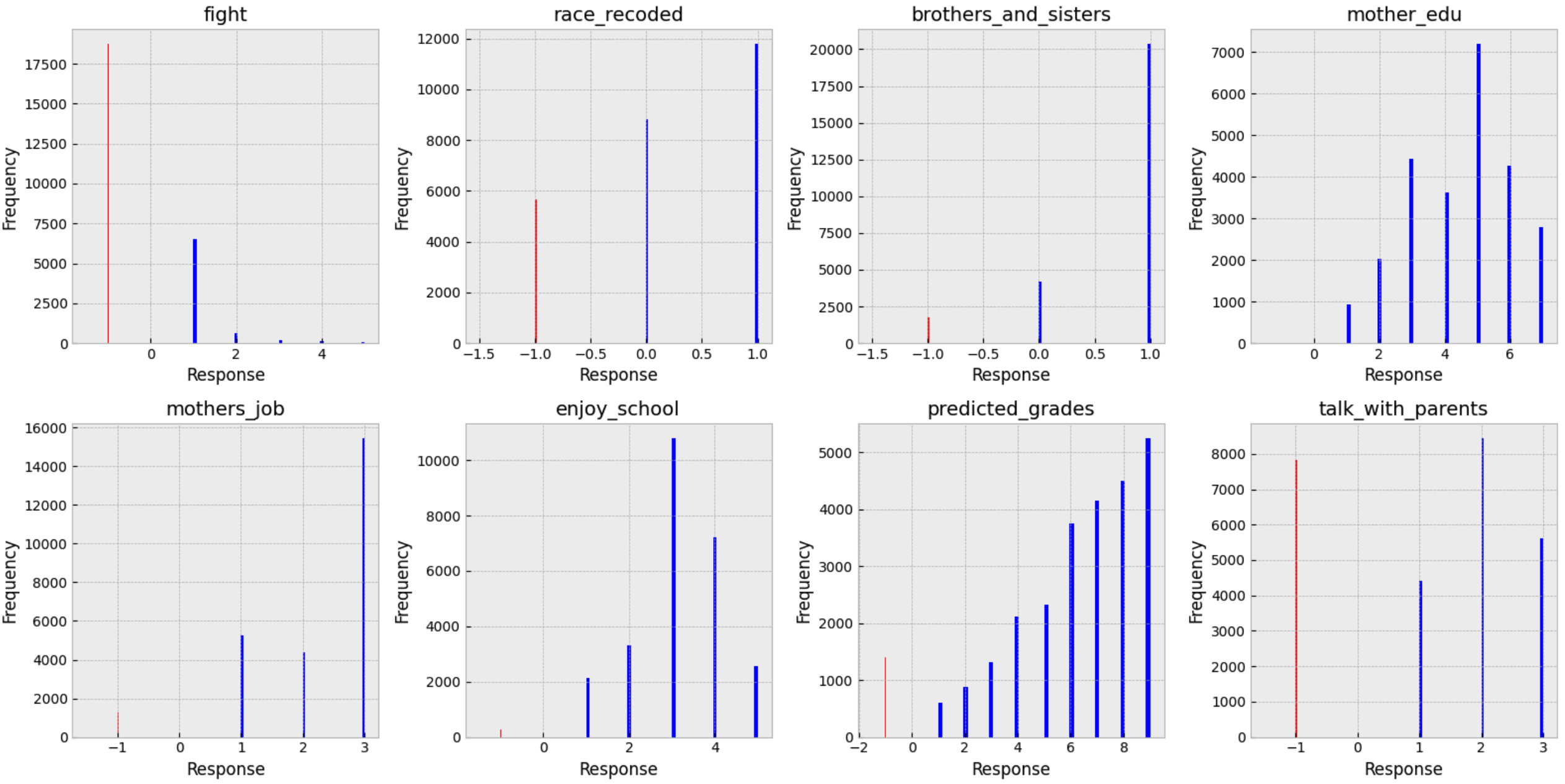}
    \caption{Missing values (\textcolor{red}{red}) and distributions (\textcolor{blue}{blue}) of the 24 different independent variables.} \vspace*{-5pt}
\end{figure}

\begin{figure}[htp]
    \centering\includegraphics[width=0.86\textwidth]{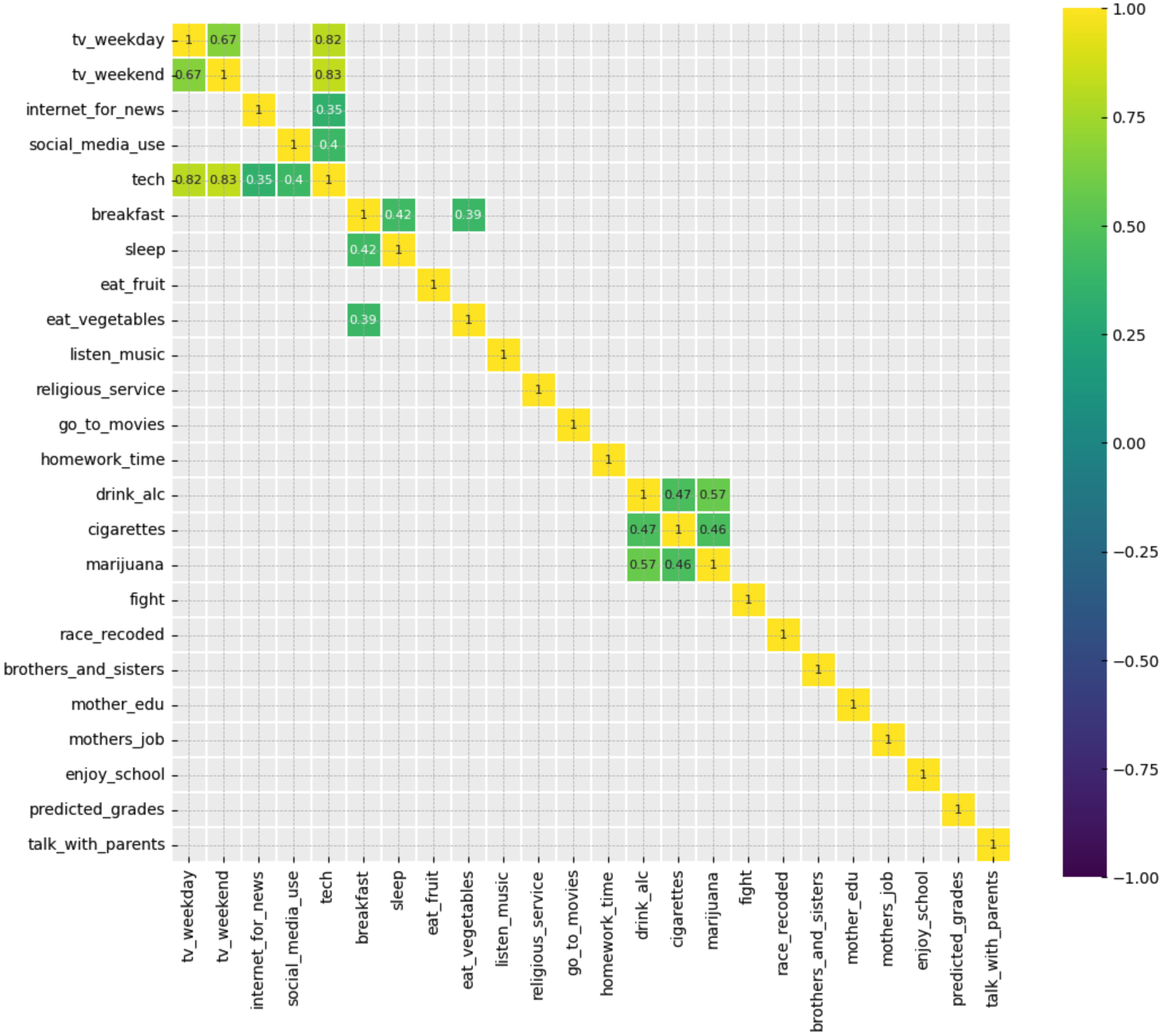}
    \caption{Correlation matrix showing the correlations between all features only if higher than $\pm0.3$.} \vspace*{-5pt}
\end{figure}

\newpage
\setcounter{table}{0}
\counterwithin{table}{section}
\section{}
\label{app:appendixD}

\begin{table}[htp]
    \caption{The registered processes that were followed and each participant's results. The asterisk (*) indicates that this paper's authors later evaluated how well the \textit{model} predicts the 20\% held-out test set based on each participant's preferred performance metrics.} \vspace*{5pt}
    \centering\includegraphics[width=0.86\textwidth]{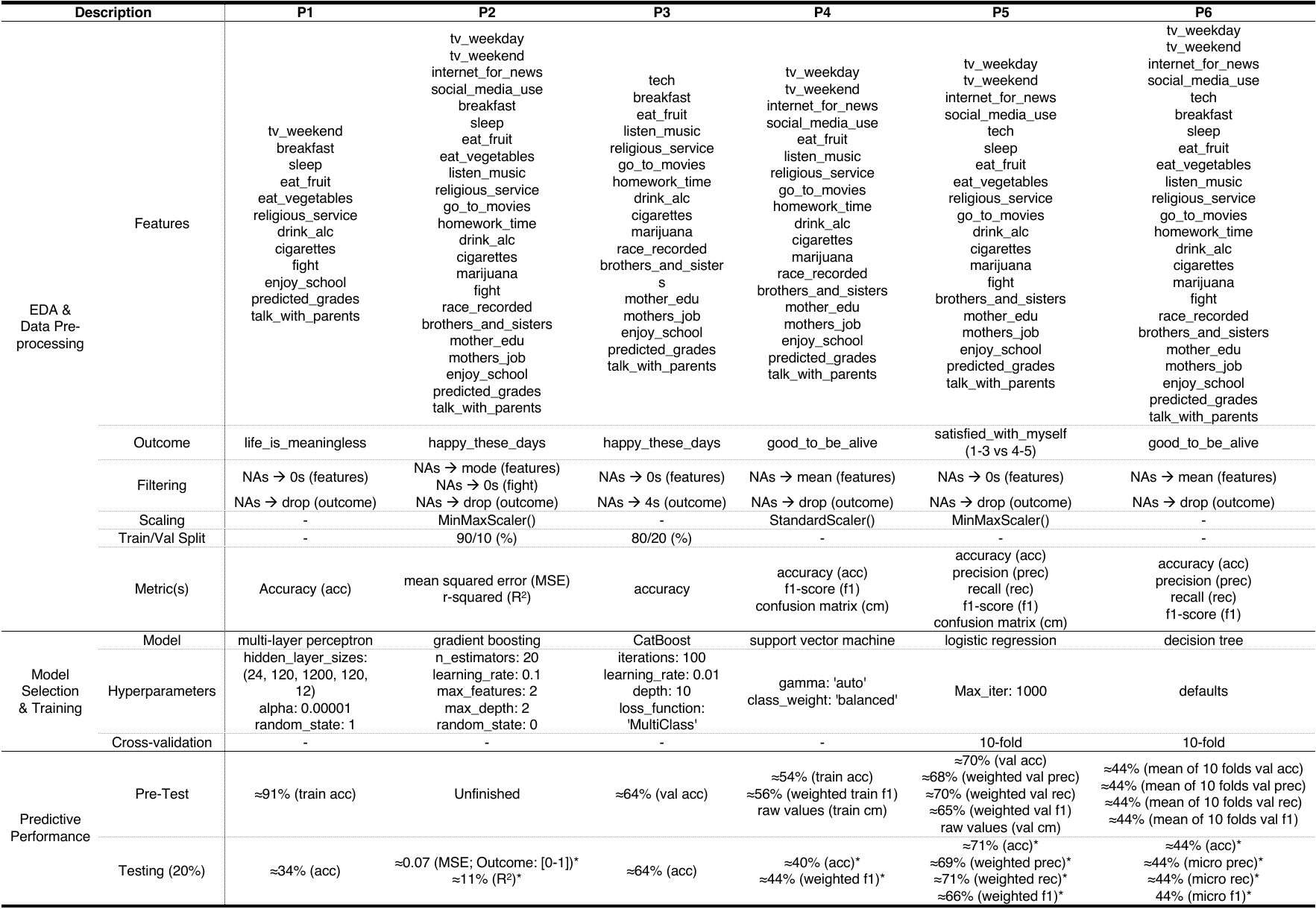}
\end{table}

\end{appendices}

\end{document}